\documentclass[lettersize,journal]{IEEEtran}
\usepackage{amsmath,amsfonts}
\usepackage{algorithmic}
\usepackage{algorithm}
\usepackage{array}
\usepackage[caption=false,font=normalsize,labelfont=sf,textfont=sf]{subfig}
\usepackage{textcomp}
\usepackage{stfloats}
\usepackage{url}
\usepackage{verbatim}
\usepackage{graphicx}
\usepackage{cite}
\usepackage{pifont}
\usepackage{booktabs}
\usepackage{multirow}
\usepackage{cuted}
\usepackage{caption}
\begin{document}

\title{SuperNeRF-GAN: A Universal 3D-Consistent Super-Resolution Framework for Efficient and Enhanced 3D-Aware Image Synthesis}

\author{Peng Zheng, Linzhi Huang, Yizhou Yu,~\IEEEmembership{Fellow,~IEEE,} Yi Chang,~\IEEEmembership{Senior Member,~IEEE,} Yilin Wang, and Rui Ma
        % <-this % stops a space
% \thanks{This paper was produced by the IEEE Publication Technology Group. They are in Piscataway, NJ. }% <-this % stops a space
\thanks{Manuscript received xxx, 2025; revised xxx, 2025. (Corresponding author: Rui Ma. Co-corresponding author: Yilin Wang.)}

\thanks{Peng Zheng  with the School of Artificial Intelligence, Jilin University, Changchun 130012, China, and also with the Shanghai Innovation Institute, Shanghai 200240, China (e-mail: zhengpeng22@mails.jlu.edu.cn).}

\thanks{Linzhi Huang, Yi Chang and Rui Ma are with the School of Artificial Intelligence, Jilin University, Changchun 130012, China (e-mail: huanglz22@mails.jlu.edu.cn; changyi@jlu.edu.cn; ruim@jlu.edu.cn).}

\thanks{Yilin Wang is with Adobe, San Jose, CA 95110, USA (e-mail: wangyilin930@gmail.com).}

\thanks{Yizhou Yu is with the Department of Computer Science, The University of Hong Kong, Pokfulam, Hong Kong (e-mail: yizhouy@acm.org).}

}

% The paper headers
\markboth{Journal of \LaTeX\ Class Files, xxx}%
{Shell \MakeLowercase{\textit{et al.}}: A Sample Article Using IEEEtran.cls for IEEE Journals}

\IEEEpubid{0000--0000/00\$00.00~\copyright~2021 IEEE}
% Remember, if you use this you must call \IEEEpubidadjcol in the second
% column for its text to clear the IEEEpubid mark.

\maketitle

\begin{abstract}
Neural volume rendering techniques, such as NeRF, have revolutionized 3D-aware image synthesis by enabling the generation of images of a single scene or object from various camera poses. However, the high computational cost of NeRF presents challenges for synthesizing high-resolution (HR) images. Most existing methods address this issue by leveraging 2D super-resolution, which compromise 3D-consistency. Other methods propose radiance manifolds or two-stage generation to achieve 3D-consistent HR synthesis, yet they are limited to specific synthesis tasks, reducing their universality.
To tackle these challenges, we propose SuperNeRF-GAN, a universal framework for 3D-consistent super-resolution. A key highlight of SuperNeRF-GAN is its seamless integration with NeRF-based 3D-aware image synthesis methods and it can simultaneously enhance the resolution of generated images while preserving 3D-consistency and reducing computational cost. Specifically, given a pre-trained generator capable of producing a NeRF representation such as tri-plane, we first perform volume rendering to obtain a low-resolution image with corresponding depth and normal map.
Then, we employ a NeRF Super-Resolution module which learns a network to obtain a high-resolution NeRF.
Next, we propose a novel Depth-Guided Rendering process which contains three simple yet effective steps, including the construction of a boundary-correct multi-depth map through depth aggregation, a normal-guided depth super-resolution and a depth-guided NeRF rendering.
Experimental results demonstrate the superior efficiency, 3D-consistency, and quality of our approach. Additionally, ablation studies confirm the effectiveness of our proposed components.
\end{abstract}


\section{Introduction}
\begin{figure*}
    \centering
    \includegraphics[width=1.0\linewidth]{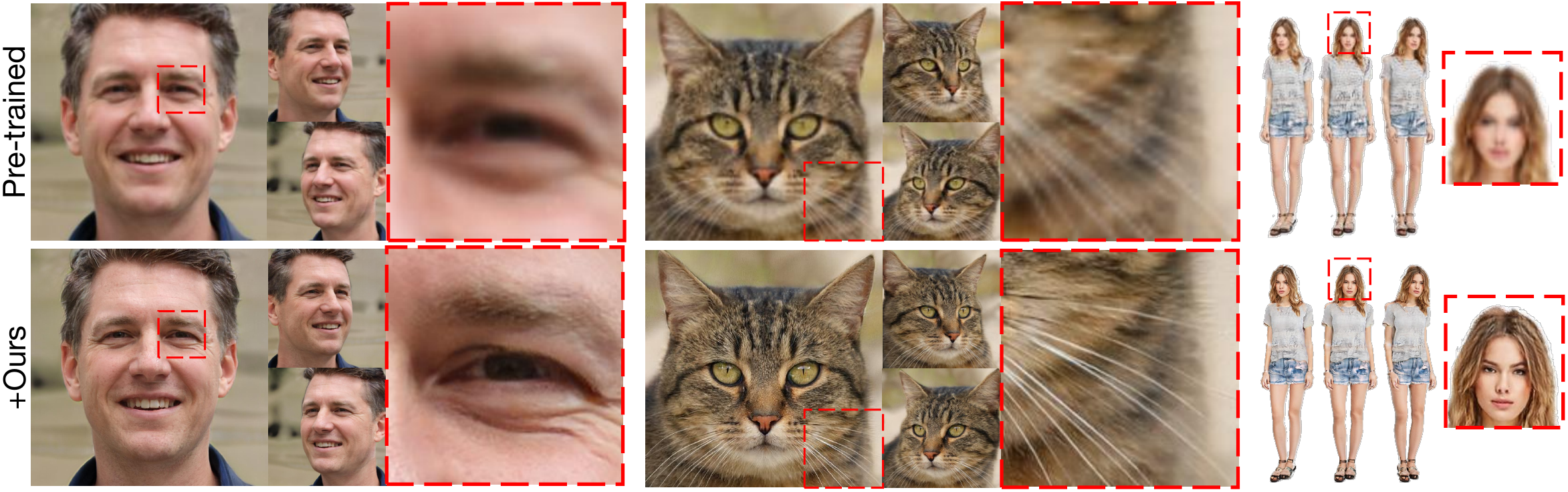}
    \caption{Effectiveness of our proposed SuperNeRF-GAN. The images in the first row are synthesized by existing pre-trained models, without the use of 2D image super-resolution. The second row shows the images super-resolved by SuperNeRF-GAN in a 3D-consistent way. Please zoom in to see the detailed differences between the original and super-resolved images. }
    \label{fig:sr}
\end{figure*}

\IEEEPARstart{T}he introduction of Neural volume rendering technique, such as NeRF~\cite{mildenhall2021nerf, barron2021mip, barron2022mip, chen2022tensorf} has significantly advanced 3D-aware image synthesis, enabling the generation of images from various camera poses for a single scene or object. These models learn NeRF representations, which can then be rendered into images at specified camera poses. However, the high computational cost inherent in NeRF limits their ability to synthesize high-resolution (HR) images. To address this, most existing methods~\cite{liu20243d, ma2023semantic, yang2023attrihuman, deng20233d, wang2023rodin} use a 2D super-resolution (SR) module, but this approach often compromises 3D-consistency. While these inconsistencies might not be evident in static images, they become apparent in free-view videos, hindering applications in areas like video games and virtual reality.

\begin{table}[t]
    \centering
    \caption{Comparison of 3D-aware image synthesis methods based on various criteria. }
    \hspace*{-0.88cm}
    \begin{tabular}{c p{0.5cm} p{0.5cm} p{0.5cm} p{0.5cm} p{0.5cm} p{0.5cm}}
    Method & \rotatebox{45}{3D-Consistent} & \rotatebox{45}{High Efficiency} & \rotatebox{45}{High-Resolution} & \rotatebox{45}{High Universality} & \rotatebox{45}{Image Quality} & \rotatebox{45}{Geometry Quality} \\
    \toprule
    StyleNeRF & \ding{56} & \ding{52} & \ding{52} & \ding{52} & \ding{56} & \ding{56} \\
    StyleSDF & \ding{56} & \ding{52} & \ding{52} & \ding{52} & \ding{56} & \ding{56} \\
    GRAM-HD & \ding{52} & \ding{52} & \ding{52} & \ding{56} & \ding{56} & \ding{56} \\
    EG3D & \ding{56} & \ding{52} & \ding{56} & \ding{52} & \ding{52} & \ding{52} \\
    SH-HD & \ding{52} & \ding{56} & \ding{52} & \ding{56} & \ding{52} & \ding{52} \\
    Ours & \ding{52} & \ding{52} & \ding{52} & \ding{52} & \ding{52} & \ding{52} \\
    \bottomrule
    \end{tabular}

    \label{tab:overall}
\end{table}

Few studies focus on 3D-consistent image synthesis. GRAM-HD~\cite{xiang2023gram} proposes using HR radiance manifolds to achieve 3D-consistent HR image synthesis. However, the inability of GRAM-HD to synthesize images from large viewpoints limits its capability for full-body and full-head image synthesis. SemanticHuman-HD (SH-HD)~\cite{zheng2024semantichuman} introduces a two-stage generation strategy for 3D-consistent HR image synthesis. However, it suffers from boundary depth issues due to its simplistic neighbor-aware depth aggregation, leading to failure in portrait synthesis. Other 3D-consistent methods struggles at HR image synthesis due to their high computational cost.

To address these limitations, we propose SuperNeRF-GAN, a universal 3D-consistent super-resolution framework that achieves HR image synthesis while maintaining 3D-consistency. This framework can be seamlessly integrated with NeRF-based 3D-aware image synthesis methods. Given a NeRF representation generated by a pre-trained model, we first perform volume rendering to obtain a low-resolution image with corresponding depth and normal maps. We then employ the NeRF Super-Resolution module, which learns a network to generate a HR NeRF representation. Following this, we introduce a novel Depth-Guided Rendering process that includes constructing a boundary-correct multi-depth map through depth aggregation and normal-guided depth super-resolution, and finally performing depth-guided NeRF rendering to synthesize HR image in a 3D-consistent way.
\IEEEpubidadjcol
% To evaluate the universality of our proposed SuperNeRF-GAN framework, we conduct comparative experiments against GRAM-HD and SH-HD, the only existing method capable of achieving 3D-consistent HR image synthesis.Our results demonstrate that SuperNeRF-GAN is the only approach that effectively handles both full-body and portrait synthesis while outperforming others in image quality. In addition, we perform further comparative experiments and ablation studies to validate the superiority and effectiveness of our method.

To evaluate the effectiveness of our method, we apply SuperNeRF-GAN to pre-trained models from state-of-the-art (SOTA) methods for portrait, cat face, and full-body image synthesis. The experimental results demonstrate significant improvements in 3D-consistency and efficiency compared to the original pre-trained models, as well as enhanced universality, 3D-consistency, and quality over other SOTA methods. 
Additionally, we conduct comparative experiments against other SOTA methods to further validate the superiority of our proposed approach. 

In summary, the main contributions of this paper are:
\begin{itemize}
    \item We propose SuperNeRF-GAN, a universal 3D-consistent super-resolution framework that enhances the resolution and 3D-consistency of synthesized images. SuperNeRF-GAN is designed to be universally applicable, making it easily deployable on NeRF-based 3D-aware image synthesis methods. 
    \item SuperNeRF-GAN overcomes the limitations of SH-HD and GRAM-HD, which are restricted to specific synthesis tasks. In contrast, SuperNeRF-GAN demonstrates high versatility, making it suitable for various synthesis tasks.
    \item Quantitative and qualitative results validate the superiority of our proposed method, particularly in terms of 3D-consistency and efficiency. 
\end{itemize}

\section{Related Work}

\subsection{3D-Aware Image Synthesis}
With the advent of generative adversarial networks (GANs)~\cite{goodfellow2014generative, karras2019style, karras2020analyzing, shi2022semanticstylegan} and diffusion models~\cite{ho2020denoising, brooks2023instructpix2pix, rombach2022high, ruiz2023dreambooth}, generative models have demonstrated impressive performance in image synthesis. Some works~\cite{tewari2020stylerig, sarkar2021style, abdal2021styleflow} achieve pose control by integrating parametric models such as SMPL~\cite{bogo2016keep} and 3DMM~\cite{egger20203d}. However, due to the lack of inherent 3D representations, these approaches do not achieve true 3D-aware image synthesis.

GRAF~\cite{schwarz2020graf} first introduces NeRF~\cite{mildenhall2021nerf} into generative models by learning a neural radiance field that can be rendered into an image at a given camera pose. However, using an MLP to model this field results in high computational cost, limiting the ability to synthesize high-quality images. Subsequent works like StyleSDF~\cite{or2022stylesdf} and StyleNeRF~\cite{gu2021stylenerf} use volume rendering to obtain a low-resolution (LR) image and then upsample it to a high-resolution (HR) image. VolumeGAN~\cite{xu20223d} employs an explicit 3D feature volume to model the radiance fields, achieving high-fidelity image synthesis. However, increasing the resolution with this 3D volume representation results in a cubic growth in computational cost, making it inefficient for HR synthesis. EG3D~\cite{chan2022efficient} addresses this by proposing a tri-plane representation, reducing the cubic growth to a quadratic level. EG3D achieves SOTA performance in both image and geometry quality, and most current 3D-aware image synthesis methods~\cite{dong2023ag3d, xu2024xagen, yang2023urbangiraffe, lei2024diffusiongan3d, an2023panohead, sun2022ide, jiang2022nerffaceediting} build upon it. However, these methods commonly employ a 2D super-resolution module, which compromises 3d-consistency although achieves HR image synthesis. As a result, these methods struggle to maintain 3d-consistency across different camera poses, limiting their application in areas such as virtual reality and video games, where both efficiency and 3d-consistency are crucial.

\begin{figure*}[t]
    \centering
    \includegraphics[width=\linewidth]{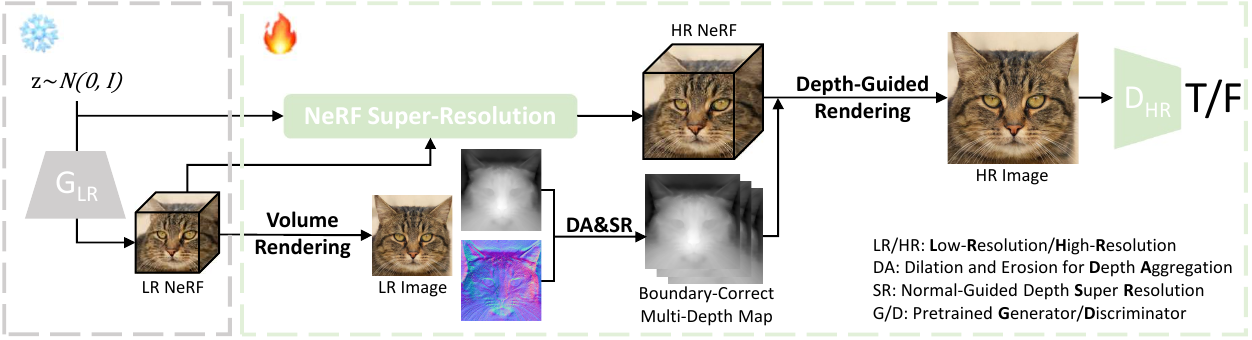}
    \caption{Pipeline of the proposed SuperNeRF-GAN framework. Given a random noise $z$, the pre-trained generator of existing 3D generative models maps it to a low-resolution (LR) NeRF representation. This LR representation can be rendered into a corresponding LR image along with depth and normal maps. Next, the LR NeRF representation undergoes the NeRF Super-Resolution module to produce a high-resolution (HR) NeRF representation. Simultaneously, Dilation and Erosion for Depth Aggregation and Normal-Guided Depth Super-Resolution are applied to the LR depth map to construct a boundary-correct multi-depth map. This map guides the rendering process of the HR NeRF representation, enabling efficient and 3D-consistent HR image synthesis.}
    \label{fig:pipeline}
\end{figure*}

\subsection{3D-Consistent HR Image Synthesis}
GRAM-HD~\cite{xiang2023gram} generates HR radiance manifolds~\cite{deng2022gram} instead of NeRF, thus avoiding the need for dense sampling and direct generation of 3D features. This approach ensures 3d-consistency by eliminating the need for image super-resolution. However, using radiance manifolds leads to suboptimal image and geometry quality. Rather than introducing a new representation like tri-plane or radiance manifolds, SemanticHuman-HD (SH-HD)~\cite{zheng2024semantichuman} proposes a two-stage generation strategy that achieves efficient 3D-consistent synthesis without compromising image quality. In the first stage, an LR image with a corresponding depth map is rendered using dense sampling. In the second stage, the depth map is aggregated using neighboring points to produce a multi-channel depth map, which can be unprojected into 3D points instead of relying on dense sampling. Although this method achieves HR image synthesis with 3d-consistency and efficiency, it suffers from boundary depth issues due to its simplistic depth aggregation.This limitation restricts SH-HD to generating depth-smooth images, such as full-body images, and prevents it from generating images in scenarios with more complex depth variations, such as portraits. As for other 3D-consistent methods~\cite{hong2022eva3d, abdal2024gaussian, chen2023veri3d}, they face challenges in HR image synthesis due to their high computational cost.

In conclusion, existing methods struggle to efficiently achieve 3D-consistent HR image synthesis for both portraits and full-body images. Our proposed SuperNeRF-GAN framework addresses these challenges by offering a universal solution that enhances both image and geometry quality while maintaining 3d-consistency across varying viewpoints.

\section{Method}
Figure~\ref{fig:pipeline} illustrates the pipeline of our SuperNeRF-GAN framework, which is designed to seamlessly integrate with existing NeRF-based 3D-aware image synthesis methods. For demonstration, we use EG3D~\cite{chan2022efficient}, a SOTA method in 3D-aware image synthesis, as a case study to introduce our SuperNeRF-GAN pipeline. Initially, our framework leverages the tri-plane representation produced by the pre-trained EG3D model to render a low-resolution image, along with associated depth and normal maps. Unlike EG3D, which performs image super-resolution in a 2D manner, our approach maintains 3D-consistency throughout the process. Specifically, we employ the NeRF Super-Resolution module and the Boundary-Correct Multi-Depth Construction technique to generate the high-resolution (HR) tri-plane representation and depth map. These components are then utilized in the Depth-Guided Rendering process, which efficiently synthesizes HR images while preserving 3D-consistency. To fully understand our methodology, we first introduce the essential preliminaries, including volume rendering~\cite{mildenhall2021nerf} and EG3D~\cite{chan2022efficient}.

\subsection{Preliminary}

\subsubsection{Volume Rendering}
NeRF~\cite{mildenhall2021nerf} introduces volume rendering to synthesize images from given camera poses. For each pixel, a ray $\textbf{r}(t)$ is cast from the camera position $\textbf{o}$ along the direction $\textbf{d}$: $\textbf{r}(t) = \textbf{o} + t\textbf{d}$, where $t$ represents the distance from the camera position. The color $C(\textbf{r})$ of this pixel is accumulated along the ray using volume rendering, which can be formulated as:
\begin{equation}
\label{eq:color}
C(\textbf{r}) = \int_{t_n}^{t_f} T(t) \cdot \sigma(\textbf{r}(t)) \cdot \textbf{c}(\textbf{r}(t)) \cdot dt,
\end{equation}
\begin{equation}
{\rm where } \ T(t) = {\rm exp} \left(-\int_{t_n}^t \sigma (\textbf{r}(s)) ds \right).
\end{equation}
Here, $\textbf{c}(\textbf{r}(t))$ and $\sigma(\textbf{r}(t))$ denote the color and density of the 3D point $\textbf{r}(t)$, respectively, while $T(t)$ represents the accumulated transmittance along the ray from $t_n$ to $t$. Notably, by replacing the color $\textbf{c}(\textbf{r}(t))$ with the normal value $\textbf{n}(\textbf{r}(t))$ and distance $t$, this formula can also be used to render the normal value $N(\textbf{r})$ and depth value $D(\textbf{r})$ of the ray. In practice, this formula is discretized. For more details about volume rendering, please refer to NeRF.

\subsubsection{EG3D as Pre-trained 3D Generator}
The generator of EG3D is adapted from StyleGAN2~\cite{karras2020analyzing}, which achieves SOTA performance in image synthesis. Given a random noise $\textbf{z}$ sampled from a Gaussian distribution, the generator maps it into a feature map with dimensions $256\times256\times96$. This feature map is then reshaped into a tri-plane representation $\textbf{T}_{LR}$, where the three planes correspond to the XY, YZ, and ZX orientations. Each plane consists of 32 channels with a resolution of $256\times256$. For a 3D point $\textbf{X}$, we project it onto these three planes and perform bilinear interpolation to extract its features from each plane. These features are then fed into an MLP, which outputs the color $\textbf{c}(\textbf{X})$ and density $\sigma(\textbf{X})$. These values are subsequently used in volume rendering, as described in Eq~\ref{eq:color}.

\subsection{3D-Consistent Low-Resolution Image Synthesis}
Using random noise $\textbf{z}$ as input, the pre-trained EG3D model generates a low-resolution tri-plane representation $\textbf{T}_{LR}$. This representation is rendered from a given camera pose to produce a low-resolution image $\textbf{I}_{LR}$, along with the corresponding depth map $\textbf{D}_{LR}$ and normal map $\textbf{N}_{LR}$, each at a resolution of $256^2$. In this process of low-resolution image synthesis, we employ dense sampling as used in EG3D. Specifically, for each pixel, 36 points are sampled along the ray using uniform sampling and an additional 36 points using importance sampling. Although dense sampling is computationally intensive, its overall computational cost remains manageable due to the low resolution of the output.

\subsection{3D-Consistent Super-Resolution}
\subsubsection{NeRF Super-Resolution}
To obtain an HR 3D representation $\textbf{T}_{HR}$, we utilize the NeRF Super-Resolution module, which is based on the architecture of StyleGAN2. This module takes the LR tri-plane representation $\textbf{T}_{LR}$ and the corresponding noise $\textbf{z}$ as input, and outputs an HR representation $\textbf{T}_{HR}$ at a resolution of $1024^2$. Importantly, the module uses the same noise $\textbf{z}$ as the pre-trained 3D generator to ensure consistency in the distribution between LR and HR images.

\subsubsection{Depth-Guided Rendering}
We construct an HR multi-depth map $\textbf{D}_{HR}$ with three channels using the Boundary-Correct Multi-Depth Map Construction, which will be detailed in the next section. This map is used to guide the sampling process, thereby avoiding the dense sampling employed by EG3D. Our Depth-Guided Rendering approach is modified from Eq.~\ref{eq:color}, and is formulated as follows:
\begin{equation}
\label{eq:depth_rendering}
    C(\textbf{r}) = \sum_{i=1}^3 T_i \left(1 - \exp(-\sigma_i \delta_i)\right) \textbf{c}_i,
\end{equation}
\begin{equation}
    {\rm where }\ T_i = \exp\left(-\sum_{j=1s}^{i-1} \sigma_j \delta_j\right).
\end{equation}
Here, \(\delta_i = \textbf{D}_{i+1} - \textbf{D}_i\) represents the distance between adjacent depth values, where \(\textbf{D}_i\) is the \(i_{th}\) value of $\textbf{D}_{HR}$. \(\sigma_i\) and \(\textbf{c}_i\) are the density and color interpolated from the HR 3D representation, with the interpolation coordinates projected from the depth map. Notably, as shown in Eq.~\ref{eq:depth_rendering}, the number of sampling points is reduced from 64 (as explained in "3D-Consistent LR Image Synthesis") to 3, achieving efficient rendering at a $1024^2$ resolution.

\begin{figure}[t]
    \centering
    \includegraphics[width=\linewidth]{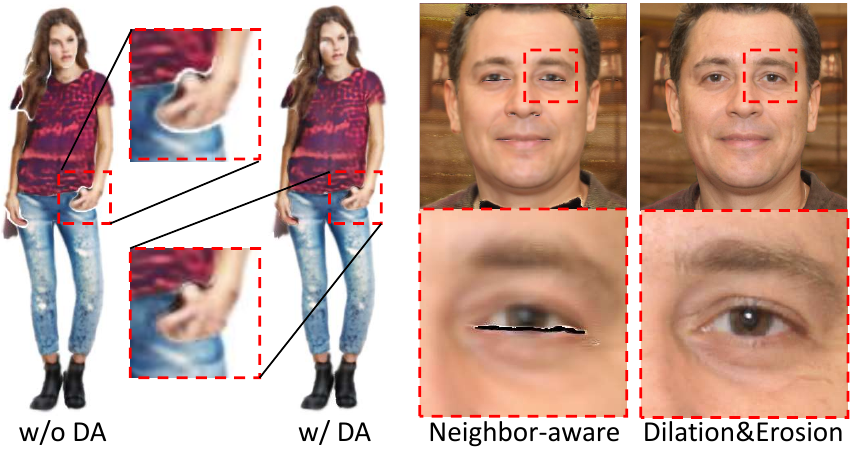}
    \caption{The left two figures demonstrate the effectiveness of Depth Aggregation (DA), note that the results are synthesized by untrained SuperNeRF-GAN models for better demonstration. The right two compare different DA techniques, highlighting that Neighbor-aware DA in SH-HD introduces noticeable artifacts, especially at depth discontinuities.}
    \label{fig:dilation}
\end{figure}
\subsection{Boundary-Correct Multi-Depth Map Construction}
\label{sec:da}
As introduced in the Depth-Guided Rendering section, we need a multi-depth map to guide the rendering. This section details how to construct this map. A straightforward approach is to perform bilinear interpolation on the low-resolution (LR) depth map $\textbf{D}_{LR}$. However, this often leads to incorrect depth values at the boundaries due to depth discontinuities. Specifically, consider two adjacent pixels on the boundary, which have non-continuous depth values. Direct interpolation will result in an averaged value that does not align with either of the two pixels' actual depth values.

To address the boundary depth issue, SH-HD~\cite{zheng2024semantichuman} proposes aggregating depth values of neighboring pixels before performing bilinear interpolation on the depth map. While this method is effective for full-body image synthesis, it struggles with portrait synthesis. This limitation arises because neighbor-aware depth aggregation does not fundamentally solve the boundary issue but merely alleviates it. Therefore, SH-HD performs well in depth-smooth scenarios, such as full-body image synthesis, but fails in other cases. To obtain a boundary-correct high-resolution depth map, our solution involves Dilation and Erosion for Depth Aggregation and Normal-Guided Depth Super-Resolution, which are introduced below.

\subsubsection{Dilation and Erosion for Depth Aggregation}
Different from SH-HD, we employ erosion and dilation operations to aggregate the depth map. Erosion is a morphological operation that shrinks the boundaries of objects in an image. For a given depth map $\textbf{D}$ and a structuring element $B$, the erosion operation is formulated as:
\begin{equation}
   (\textbf{D} \ominus B)(x,y) = \min_{(s,t) \in B} \{ \textbf{D}(x+s, y+t) - B(s,t) \},
\end{equation}
where \( (x,y) \) are the coordinates of a pixel in the depth map \textbf{D}, and \( (s,t) \) are the coordinates within the structuring element $B$. This operation slides the structuring element $B$ over the image \textbf{D} and replaces each pixel by the minimum value of the image pixels covered by $B$ minus the corresponding value of $B$. Conversely, dilation expands the boundaries of objects in an image. The dilation operation is formulated as:
\begin{equation}
   (\textbf{D} \oplus B)(x,y) = \max_{(s,t) \in B} \{ \textbf{D}(x+s, y+t) + B(s,t) \}.
\end{equation}
The choice of the structuring element $B$ significantly affects the results of erosion and dilation. In our case, we use a square structuring element. By applying erosion and dilation to the LR depth map $\textbf{D}_{LR}$, we obtain the dilated depth map $\textbf{D}_{dil}$ and eroded depth map $\textbf{D}_{ero}$, which are then concatenated with $\textbf{D}_{LR}$, resulting in an aggregated map $\textbf{D}_{agg}$ with three channels. This aggregated map addresses the boundary depth issue by storing depth values from both sides of the boundary, as shown in Figure~\ref{fig:dilation} (right).

% \begin{figure}[t]
%     \centering
%     \includegraphics[width=\linewidth]{figure/dilation.pdf}
%     \caption{Visualization of dilation and erosion operations. The structuring element used has a large kernel size for better visualization, which differs from the one used in our framework.}
%     \label{fig:dilation}
% \end{figure}

\begin{figure}
    \centering
    \includegraphics[width=\linewidth]{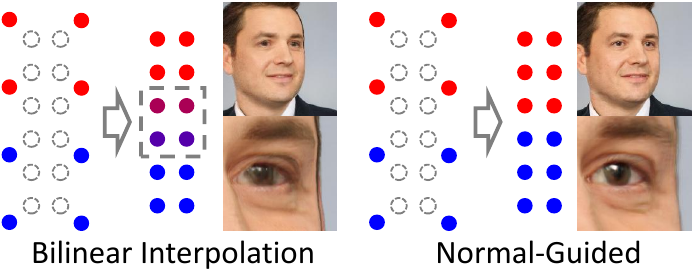}
    \caption{Effectiveness of Normal-Guided Depth Super-Resolution. The dashed rectangle highlights inaccuracies at depth discontinuity using bilinear interpolation, which result in artifacts as in the synthesized image.}
    \label{fig:normal}
\end{figure}

\begin{figure*}[t]
    \centering
    \includegraphics[width=\textwidth]{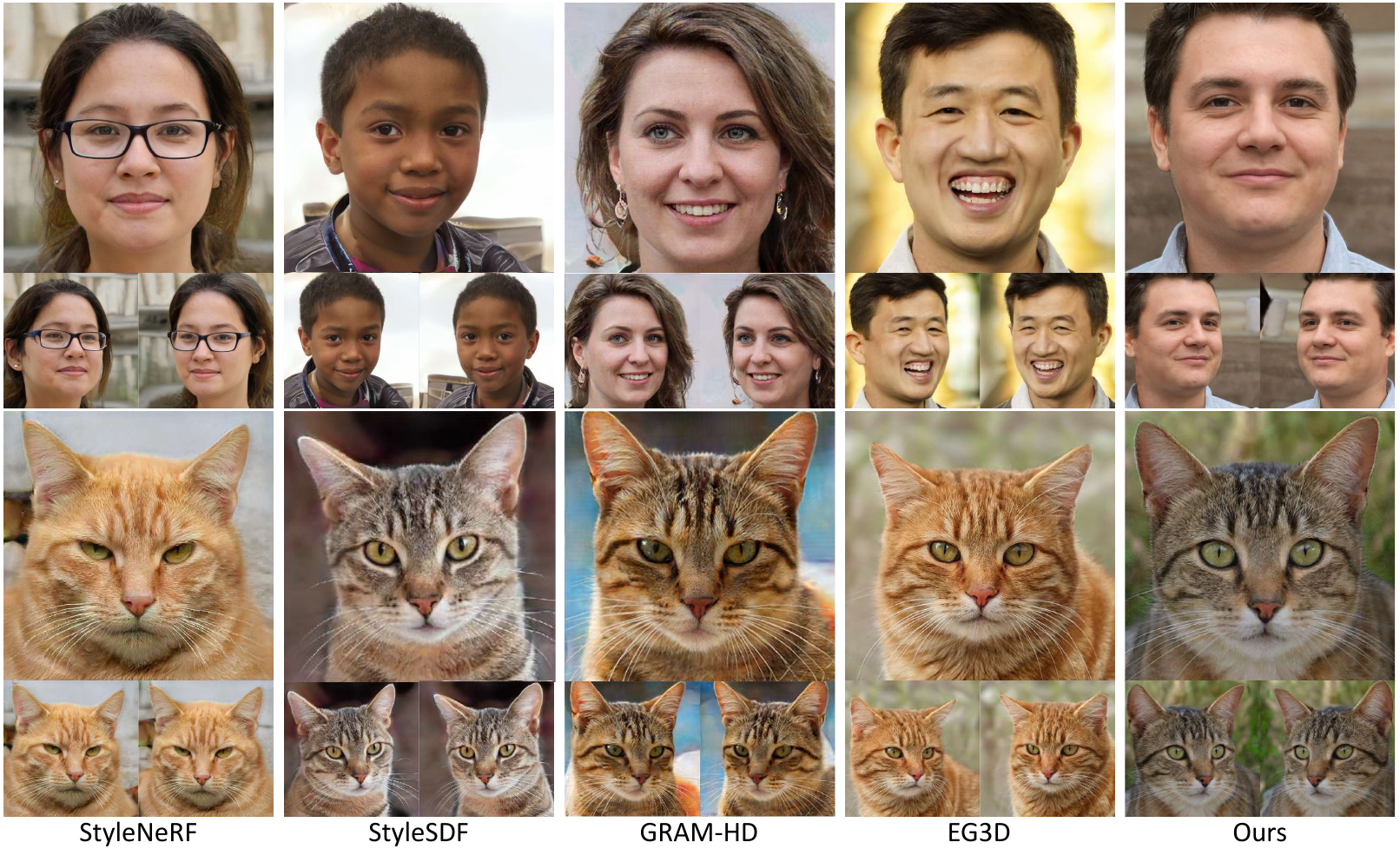}
    \caption{Qualitative comparison among 3D-aware image synthesis methods. The results of other methods are taken from their respective papers to ensure a fair and consistent comparison. Since the 3D-inconsistency might not be evident in static images, we provide additional comparisons of 3D-consistency in our \textbf{Supplementary Video}, where StyleNeRF, StyleSDF, and EG3D show noticeable inconsistencies.}
    \label{fig:3d}
\end{figure*}

\subsubsection{Normal-Guided Depth Super-Resolution}
\label{sec:normal}
Naive bilinear interpolation on the aggregated depth map results in multi depth values at boundaries, which include values from both sides of the boundary. However, these depth values may not be accurate as the bilinear interpolation relies solely on the information within the aggregated depth map. To address this, we propose a Normal-Guided Depth Super-Resolution module, which leverages the normal map to provide supplementary geometric information.

Given a normal map $\textbf{N} = (N_x, N_y, N_z)$, the differences in depth values can be computed as: $\Delta_x = N_x / N_z$ and $\Delta_y = N_y / N_z$. The super-resolved depth map $\textbf{D}_{SR}$ at $512^2$ resolution can be expressed as the original depth map plus the differences with respect to $x$ and $y$, formulated as:
\begin{equation}
\textbf{D}_{SR}(x, y) = \textbf{D}_{LR}(m, n) + \Delta(x,y),
\end{equation}
\begin{equation}
    \Delta(x,y) = \frac{w_x(x,y) \cdot \Delta_x(m, n) + w_y(x,y) \cdot \Delta_y(m, n)}{\sqrt{(\Delta_x(m, n))^2 + (\Delta_y(m, n))^2}},
\end{equation}
where \(w_x(x, y)\) and \(w_y(x, y)\) are weight functions dependent on $x$ and $y$. The weight function is defined as:
\begin{equation}
w_x(x,y) = \frac{e^{-\lvert\Delta_x(m, n)\rvert}}{e^{-\lvert\Delta_x(m, n)\rvert} + e^{-\lvert\Delta_y(m, n)\rvert}} \cdot (-1)^{\mathbb{I}(x = 2m - 1)},
\end{equation}
where $\mathbb{I}$ is an indicator function while \(m\) and \(n\) are integers related to the coordinates \(x\) and \(y\) by:
\begin{equation}
    m = \left\lfloor \frac{x + 1}{2} \right\rfloor, \quad n = \left\lfloor \frac{y + 1}{2} \right\rfloor.
\end{equation}
The remaining part of the weight function is a softmax function inspired by BiNi~\cite{cao2022bilateral}, which assumes that the depth map is semi-smooth. This means that the depth map is one-sided differentiable, with larger weight assigned to the direction with smaller differences.

Consequently, the aggregated depth map $\textbf{D}_{agg}$ is processed through the Normal-Guided Depth Super-Resolution module twice, resulting in a boundary-correct multi-depth map $\textbf{D}_{HR}$ at $1024^2$ resolution. Notably, to facilitate Depth-Guided Rendering, the three-channel depth values of this multi-depth map are sorted. A visualization of this module is provided in Figure~\ref{fig:normal}.

\subsection{Training Pipeline}

\subsubsection{Training Strategy}
Our method can be directly applied to pre-trained 3D generative models, such as EG3D and SH-HD. During training, we freeze the pre-trained generator and focus on training the NeRF Super-Resolution module. Notably, the discriminator has been redesigned to process high-resolution images and is tunable during training.

\subsubsection{Loss Function}
We use the loss functions from the original method to ensure training consistency. Additionally, we adopt the unsample loss from SH-HD to guide the training of the NeRF Super-Resolution module by penalizing inconsistency between high-resolution and low-resolution images.

%Additionally, we propose a progressive SR loss to guide the training. Specifically, we progressively decrease the weight of SR loss. Consequently, the weight in early stage is bigger to guide the SR module, ensuring the consistency between images at high resolution and low resolution. In the later stage, the small weight of SR loss allow the SR module to learn more image details.

\section{Implementation Details}
\subsubsection{Training Setup}
Our experiments were conducted on a server equipped with 4 NVIDIA A40 GPUs, each with 48GB of memory. The models are trained over a period of 2 to 4 days, depending on the complexity and size of the dataset. Specifically, for portrait and cat face image synthesis tasks, we used a batch size of 16, while a batch size of 4 was employed for full-body image synthesis due to the increased computational demands.

\subsubsection{SH-HD for Portrait and Cat Face Image Synthesis}
To facilitate portrait and cat face image synthesis, we integrate key components from SH-HD~\cite{zheng2024semantichuman} into pre-trained EG3D models. These components include the two-stage generation strategy, feature super-resolution module, unsample loss, and neighborhood-aware depth aggregation technique. Despite these adaptations, SH-HD exhibits limitations in portrait and cat face image synthesis, resulting in a drop in image quality for these specific tasks, as demonstrated in Fig.~\ref{fig:dilation}. The results underscore the superior universality of our proposed method.

\section{Experiments}
\subsubsection{Datasets}
We train our models on three distinct datasets for specific synthesis tasks: FFHQ~\cite{karras2019style}, which includes 50K portrait images at a resolution of $1024^2$, is used for portrait synthesis. For cat face synthesis, we use the AFHQ~\cite{choi2020stargan}, which contains 5.5K cat face images at a resolution of $1024^2$. For full-body image synthesis, we employ DeepFashion~\cite{liu2016deepfashion}, which provides 7K full-body images at a resolution of $1024^2$, along with corresponding segmentation maps. Additionally, due to the requirement in SH-HD~\cite{zheng2024semantichuman}, we also incorporate normal maps from SH-HD and human poses from AG3D~\cite{dong2023ag3d}.

\subsubsection{Baselines}
For portrait and cat face image synthesis, our model leverages pre-trained EG3D~\cite{chan2022efficient} models. We compare our results with GRAM-HD~\cite{xiang2023gram}, StyleNeRF~\cite{gu2021stylenerf}, and StyleSDF~\cite{or2022stylesdf}, all recognized for their high-resolution image synthesis capabilities. However, StyleNeRF and StyleSDF rely on 2D super-resolution, compromising the 3D consistency of the synthesized images. In contrast, GRAM-HD's manifold representation limits image quality, geometry accuracy, and universality. For full-body image synthesis, our model builds on the pre-trained SH-HD model. We compare it with EVA3D, VeRi3D, and GSM, which are noted for their 3D-consistent image synthesis. Nevertheless, their reliance on vertex-based, MLP-based, or Gaussian shell-based representations imposes limitations on the quality of the generated images.

\subsubsection{Evaluation Metrics}
We use Frechet Inception Distance (FID)~\cite{heusel2017gans} and Kernel Inception Distance (KID)~\cite{binkowski2018demystifying} to assess the quality of synthesized images. Note that all KID scores are multiplied by 1000. To evaluate 3D-consistency across images synthesized from different camera poses, we adopt Peak Signal-to-Noise Ratio (PSNR) and Structural Similarity (SSIM) metrics, as used in GRAM-HD. Higher scores indicate better 3D-consistency. Specifically, we reconstruct a NeuS~\cite{wang2021neus} model from images synthesized from various camera angles and calculate the PSNR and SSIM scores between the synthesized and reconstructed images. These scores are averaged across 50 entities. For each entity, we generate images from 30 uniformly sampled yaw angles ranging from -0.4 to 0.4 radians. These images are subsequently used to construct a NeuS~\cite{wang2021neus} representation, utilizing NeuS's default settings to ensure consistency and comparability in our experiments. The constructed NeuS representation is subsequently rendered into reconstructed images from the specified angles to facilitate evaluation.

\begin{table}[t]
    \centering
    % \caption{Ablation study of our proposed SuperNeRF-GAN. Incorporating SuperNeRF-GAN consistently improves 3D-consistency. Although there is a slight degradation in image quality, we attribute this to the 2D SR network used in EG3D, which enhances image quality at the cost of 3D-consistency. In contrast, our method maintains complete 3D-consistency and supports high-resolution image synthesis at $1024^2$. Note that all KID scores in this paper are multiplied by 1000.}
    \caption{Effectiveness of SuperNeRF-GAN for portrait image synthesis. Note that all KID scores in this paper are multiplied by 1000.}
    \tabcolsep=0.17cm
    \begin{tabular}{c c c c c c c c}
    \toprule
    Dataset & Method & FID$\downarrow$ & KID$\downarrow$ & PSNR$\uparrow$ & SSIM$\uparrow$ & Res \\
    \midrule
    \multirow{3}*{FFHQ} & EG3D & \pmb{4.65} & \pmb{1.27} & 33.67 & 0.893 & 512\\
    \cmidrule{2-7}
    ~ & \multirow{2}*{+SuperNeRF-GAN} & 5.13 & 1.70 & 34.36 & 0.920 & 512\\
    ~ & ~ & 5.10 & 1.54 & \pmb{36.44} & \pmb{0.935} & 1024\\
    \midrule
    \multirow{2}*{AFHQ} & EG3D & \pmb{3.19} & \pmb{0.38} & 32.61 & 0.843 & 512\\
    % \cmidrule{2-7}
    ~ & +SuperNeRF-GAN & 3.77 & 1.09 & \pmb{33.01} & \pmb{0.861} & 512\\
    \bottomrule
    \end{tabular}
    \label{tab:eg3d}
\end{table}

\begin{table}[t]
\caption{Quantitative comparison with SH-HD. "Mem" indicates the GPU memory consumption during training. Since SH-HD is not directly applicable to the FFHQ and AFHQ datasets, we modified it to enable training on these datasets.}
\tabcolsep=0.12cm
\begin{tabular}{c c c c c c c c c c }
\toprule
\multirow{2}*{Method} & \multicolumn{3}{c}{DeepFashion1024} & \multicolumn{3}{c}{FFHQ1024} & \multicolumn{3}{c}{AFHQ512} \\
\cmidrule{2-10}
~ & FID$\downarrow$ & KID$\downarrow$ & Mem & FID$\downarrow$ & KID$\downarrow$ & Mem & FID$\downarrow$ & KID$\downarrow$ & Mem \\
\midrule
SH-HD & 8.70 & 4.04 & 31G & 31.9 & 27.8 & 25G & 8.77 & 11.7 & 12G \\
Ours & \pmb{8.47} & \pmb{3.68} & \pmb{14G} & \pmb{5.10} & \pmb{1.54} & \pmb{11G} & \pmb{3.77} & \pmb{1.09} & \pmb{8G} \\
\bottomrule
\label{tab:sh-hd}
\end{tabular}
\end{table}

\subsection{Comparisons}
We evaluate 3D-aware image synthesis methods across various criteria, with the results presented in Table~\ref{tab:overall}. Note that only our method achieves 3D-consistent HR image synthesis with high universality. Specifically, GRAM-HD is unable to synthesize full-body images due to inherent limitations in its manifolds, and SH-HD struggles with generating high-quality portraits because of boundary issues, which restrict their universality. 

As our models are deployed on the pre-trained 3D generators of EG3D and SH-HD, we compare our approach with these methods, as illustrated in Table~\ref{tab:eg3d} and \ref{tab:sh-hd}. Compared to EG3D, our method shows stable improvements in 3D-consistency, albeit with a slight compromise in image quality. We argue that the 2D super resolution used in EG3D significantly improve the image quality, whereas SuperNeRF-GAN leverages Depth-Guided Rendering to ensure 3D-consistency. It is worth noting that the quantitative improvements in 3D-consistency may appear subtle, as they are constrained by the limitations of the reconstruction method NeuS. We encourage readers to refer to the \textbf{Supplementary Video} for more qualitative comparisons, where the advantages of our method are clearly evident. As for SH-HD, which also ensures 3D-consistency, our method outperforms it in both efficiency and image quality. Notably, SH-HD performances poorly on the FFHQ and AFHQ datasets due to its naive depth aggregation technique.  

\subsubsection{3D-Consistency}

\begin{table}[t]
\centering
\caption{Quantitative comparison of 3D-consistency. $^\ast$Results taken from GRAM-HD. }
\tabcolsep=0.35cm
    \begin{tabular}{c c c c c}
    \toprule
    \multirow{2}*{Method} & \multicolumn{2}{c}{FFHQ1024} & \multicolumn{2}{c}{AFHQ512} \\
    \cmidrule{2-5}
    ~ & PSNR$\uparrow$ & SSIM$\uparrow$ & PSNR$\uparrow$ & SSIM$\uparrow$ \\
    \midrule
    StyleNeRF$^\ast$ & 30.0 & 0.804 & - & - \\
    StyleSDF$^\ast$ & 31.1 & 0.836 & 26.6 & 0.749 \\
    GRAM-HD$^\ast$ & 33.8 & 0.872 & 28.8 & 0.807 \\
    Ours & \pmb{36.4} & \pmb{0.935} & \pmb{33.0} & \pmb{0.861} \\
    \bottomrule
    \end{tabular}
    \label{tab:3d}
\end{table}

\begin{table}[t]
    \centering
    \caption{Quantitative comparison of image quality among methods that achieve high-resolution ($1024\times1024$) image synthesis. $^\ast$Results are taken from GRAM-HD. }
    \tabcolsep=0.46cm
    \begin{tabular}{c c c c c c}
    \toprule
    \multirow{2}*{Method} & \multicolumn{2}{c}{FFHQ1024} & \multicolumn{2}{c}{AFHQ512} \\
    \cmidrule{2-5}
    ~ & FID$\downarrow$ & KID$\downarrow$ & FID$\downarrow$ & KID$\downarrow$ \\
    \midrule
    StyleNeRF$^\ast$ & 9.45 & 2.65 & - & - \\
    StyleSDF$^\ast$ & 9.44 & 2.83 & 7.91 & 3.90 \\
    GRAM-HD$^\ast$ & 12.0 & 5.23 & 7.67 & 3.41 \\
    Ours & \pmb{5.10} & \pmb{1.54} & \pmb{3.77} & \pmb{1.09} \\
    \bottomrule
    \end{tabular}
    \label{tab:fid}
\end{table}

\begin{table}[t]
    \centering
    \caption{Quantitative comparison among methods that achieve full-body image synthesis with 3D-consistency. $^\ast$Results taken from their papers.}
    \tabcolsep=0.4cm
    \begin{tabular}{c c c c c}
    \toprule
    Method & FID$\downarrow$ & KID$\downarrow$ & Resolution & Memory \\
    \midrule
    EVA3D & 15.89 & 9.25 & 512 & 33G \\
    GSM & 15.78$^\ast$ & - & 512 & - \\
    VeRi3D & 21.4$^\ast$ & - & 512 & 34G \\
    SH-HD & 8.70 & 4.04 & 1024 & 31G \\
    \midrule
    \multirow{2}*{Ours} & 10.56 & 5.60 & 512 & 11G \\
    ~ & \pmb{8.47} & \pmb{3.68} & 1024 & 14G \\
    \bottomrule
    \end{tabular}
    \label{tab:human}
\end{table}

The comparison of 3D-consistency is shown in Table~\ref{tab:3d}. StyleNeRF and StyleSDF exhibit poorer 3D-consistency due to their 2D super-resolution module. While GRAM-HD performs better than the aforementioned methods, its radiance manifolds constrain image quality. Consequently, our method consistently delivers superior performance in both 3D-consistency and image quality. We also present additional qualitative comparison between various 3D-aware image synthesis methods, as illustrated in Fig.~\ref{fig:3d}. The results from other methods were directly taken from their respective publications to ensure a fair and consistent comparison. Since 3D-inconsistency might not be evident in static images, we provide additional comparisons of 3D-consistency in our \textbf{Supplementary Video}, where StyleNeRF, StyleSDF, and EG3D show noticeable inconsistencies. In contrast, both GRAM-HD and our method achieve 3D-consistent synthesis. However, GRAM-HD introduces artifacts when viewed from large angles, where our method performs well. 

\begin{figure*}
    \centering
    \includegraphics[width=\linewidth]{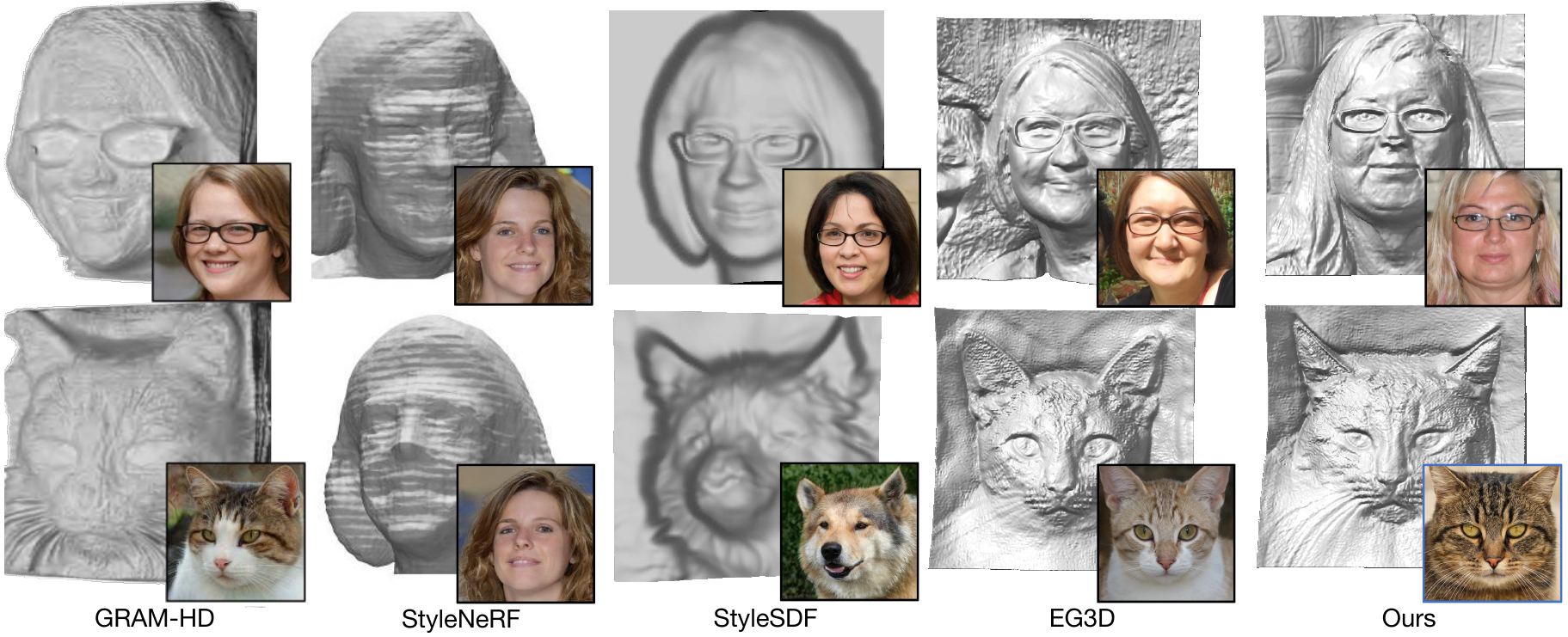}
    \caption{Qualitative comparison of geometry quality. EG3D and our method generate results with more detailed geometry than other methods. The results for GRAM-HD, StyleNeRF, and StyleSDF are directly taken from their respective papers.}
    \label{fig:geo}
\end{figure*}

\subsubsection{Image Quality}
In terms of image quality, our method shows significant improvements, as shown in Table~\ref{tab:fid}. Notably, although GRAM-HD employs generative radiance manifolds to achieve 3D-consistent image synthesis, it compromises on image quality compared to super-resolution-based methods such as StyleNeRF and StyleSDF. This trade-off highlights that achieving strong 3D-consistency imposes stronger constraints on generators, often leading to a degradation in image quality. Consequently, we argue that the slight compromise in image quality in our method, as compared to EG3D (Table~\ref{tab:eg3d}), is acceptable. On the other hand, while our method prioritizes 3D-consistency, it still achieves better image quality than StyleNeRF and StyleSDF, and is comparable to EG3D. Furthermore, we evaluated the image quality of 3D-consistent methods on full-body image synthesis task, with results presented in Table~\ref{tab:human}. Our method not only achieves the highest image quality but also supports high-resolution image synthesis with low memory consumption, thanks to our efficient 3D-consistent super-resolution module.

\subsubsection{Efficiency}
Our method significantly enhances efficiency in high-resolution image synthesis due to its Depth-Guided Rendering approach, which reduces the number of sampling points. To assess the efficiency of our method, we perform comparative experiments with other 3D-consistent image synthesis methods. As shown in Table~\ref{tab:human}, our method offers substantial improvements in efficiency, achieving $1024^2$ resolution image synthesis with only 14G of GPU memory.

\subsubsection{Universality}
As noted in the GRAM-HD paper, this method cannot synthesize images from wide viewpoints due to its reliance on the near-plane manifold representation for efficient rendering. In contrast, our method leverages depth-guided rendering, which is compatible with any NeRF representation and enables the synthesis of images from wide viewpoints, such as full-body images. Another 3D-consistent method, SH-HD, struggles with unsmooth depth values, as shown in Table~\ref{tab:sh-hd}, due to its naive neighbor-aware depth aggregation technique. This issue is further evidenced in Fig.\ref{fig:dilation}. Consequently, our method stands as the only 3D-consistent image synthesis approach with robust universality, as demonstrated in Table\ref{tab:overall}.

\subsubsection{Geometry Quality}

We conduct a qualitative comparison of geometry quality, as shown in Fig.~\ref{fig:geo}. GRAM-HD employs radiance manifolds to achieve efficient rendering, with each manifold being nearly planar. As a result, the generated geometries are essentially the combination of several "planes," which limits the overall quality. StyleSDF and StyleNeRF generate NeRF representations at relatively low resolutions (e.g., 32), which further constrains their ability to capture fine geometric details. In contrast, our method produces high-resolution NeRF representations, achieving the same geometry quality as EG3D.

% \begin{figure*}[t]
%   \centering
%   %   \begin{minipage}[t]{\textwidth}
%   %     \centering
%   %   \includegraphics[width=\linewidth]{figure/pose.pdf}
%   %   \caption{3D-consistent image synthesis. The images synthesized from different camera poses exhibit consistent 3D structures.}
%   %   \label{fig:pose}
%   % \end{minipage}
%   \begin{minipage}[t]{\textwidth} 
%       \centering
%     \includegraphics[width=\linewidth]{figure/inv.pdf}
%     \caption{Embedding of images at a resolution of $1024^2$ using GAN inversion technique. }
%     \label{fig:inv}
%   \end{minipage}
% \end{figure*}

\begin{table}[t]
    \centering
    \caption{Ablation study on depth aggregation methods. "D\&E" denotes the dilation and erosion operations used in our method, while (D\&E)$^2$ signifies performing dilation and erosion twice to yield 5 depth values per pixel.}
    \tabcolsep=0.27cm
\begin{tabular}{c c c c c c c}
\toprule
\multirow{2}*{Method} & \multicolumn{2}{c}{DeepFashion} & \multicolumn{2}{c}{FFHQ1024} & \multicolumn{2}{c}{AFHQ512} \\
\cmidrule{2-7}
~ & FID$\downarrow$ & KID$\downarrow$ & FID$\downarrow$ & KID$\downarrow$ & FID$\downarrow$ & KID$\downarrow$ \\
\midrule
(D\&E)$^2$ & 9.10 & 4.71 & 6.47 & 2.95 & 4.50 & 1.27 \\
D\&E & \pmb{8.47} & \pmb{3.68} & \pmb{5.10} & \pmb{1.54} & \pmb{3.77} & \pmb{1.09}\\
\bottomrule
\end{tabular}
\label{tab:aggregation}
\end{table}

\subsection{Ablation Study}
\subsubsection{Depth Aggregation}
% \begin{table*}[]
%     \centering
%     \begin{tabular}{c c c c c c c c c c }
%     \toprule
%     Method & FID$\downarrow$ & KID$\downarrow$ & Memory & FID$\downarrow$ & KID$\downarrow$ & Memory & FID$\downarrow$ & KID$\downarrow$ & Memory \\
%     11 Points & 8.70 & 4.04 & 31G & 31.9 & 27.8 & 25G & 8.77 & 11.7 & 12G
%     \end{tabular}
%     \caption{Caption}
%     \label{tab:my_label}
% \end{table*}

The results in Fig.\ref{fig:dilation} confirm the effectiveness of our depth aggregation technique, as discussed in Section\ref{sec:da}. Specifically, without depth aggregation, noticeable holes appear in the generated images due to the inaccuracies caused by straightforward bilinear interpolation. In contrast, the naive neighbor-aware depth aggregation proposed in SH-HD suffers from artifacts at depth discontinuities. This approach only partially mitigates the boundary issue without fundamentally resolving it. This claim is further supported quantitatively in Table~\ref{tab:sh-hd}, where our method demonstrates greater universality compared to SH-HD, primarily due to the differences in depth aggregation techniques.

To further investigate depth aggregation, we conduct a quantitative ablation study, as presented in Table~\ref{tab:aggregation}. In this study, we apply additional dilation and erosion operations to produce a 5-channel aggregated depth map. The results show that doubling the depth aggregation does not improve performance and instead introduces depth inaccuracies. While our Normal-Guided Depth Super-Resolution mitigates these inaccuracies, the additional depth values fail to yield benefits.

\begin{table}[t]
    \centering
    \caption{Ablation study on interpolation methods. Our proposed Normal-Guided Super-Resolution achieve improvements on 3D-consistency. The models are trained on the FFHQ1024 dataset, as high resolution better highlights these improvements. }
    \tabcolsep=0.33cm
\begin{tabular}{c c c c c c}
\toprule
Method & FID$\downarrow$ & KID$\downarrow$ & PSNR$\uparrow$ & SSIM$\uparrow$ \\
\midrule
Bilinear Interpolation & \pmb{5.08} & 1.51 & 35.75 & 0.915 \\
Normal-Guided & \pmb{5.08} & \pmb{1.48} & \pmb{36.44} & \pmb{0.935} \\
\bottomrule
\end{tabular}
\label{tab:interpolation}
\end{table}

\subsubsection{Normal-Guided Depth Super-Resolution}
To evaluate the effectiveness of our proposed Normal-Guided Depth Super-Resolution module, we perform a quantitative comparison against bilinear interpolation. As shown in Table~\ref{tab:interpolation}, our method achieves notable improvements in 3D-consistency. These improvements stem from the module's ability to accurately super-resolve depth maps with guidance from the normal map. As illustrated in Fig.\ref{fig:normal}, compared to naive bilinear interpolation, our normal-guided approach leverages the normal map to identify and handle depth discontinuities at boundaries more effectively. This capability ensures smoother and more accurate depth transitions. For a detailed explanation, please refer to Section\ref{sec:normal}.

\begin{figure*}
    \centering
    \includegraphics[width=\linewidth]{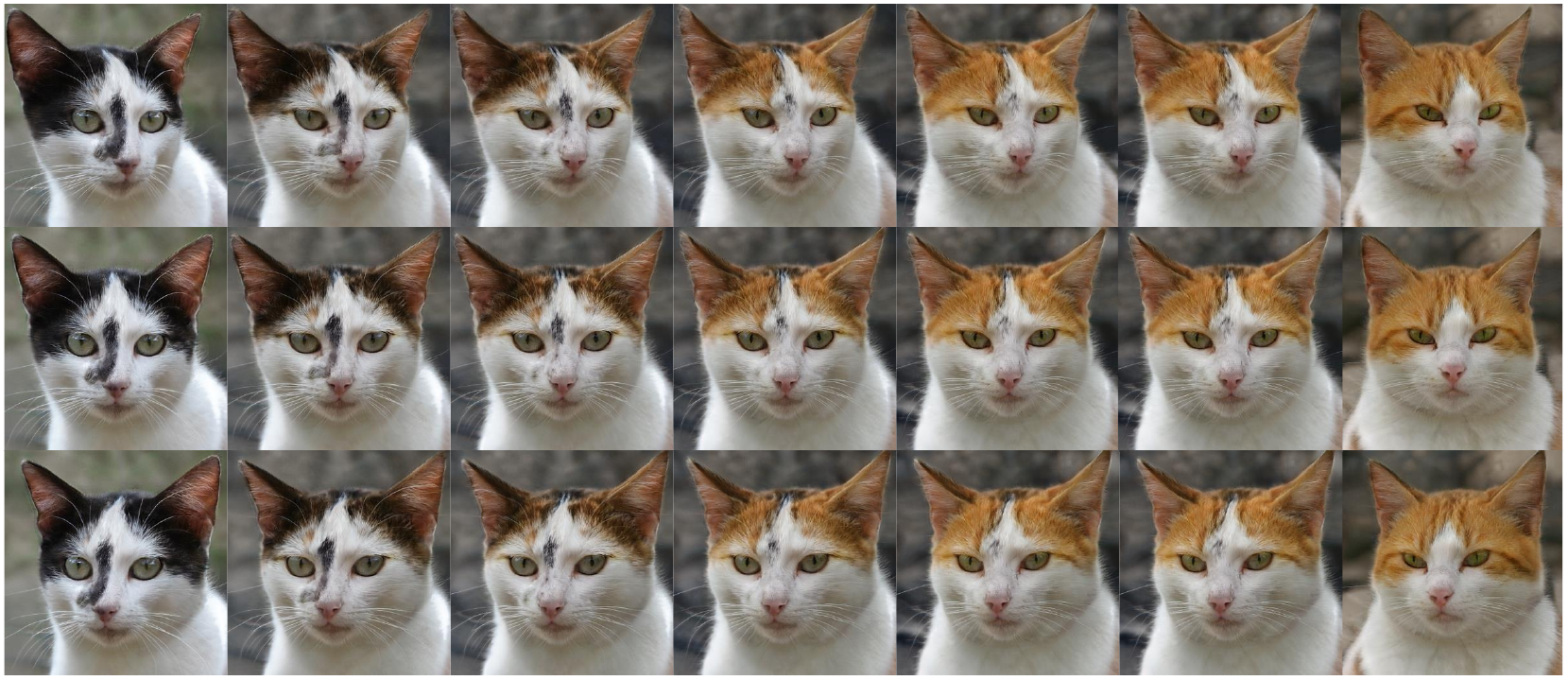}
    \caption{Interpolation in the latent space produces smooth transitions, with each intermediate result maintaining 3D-consistency.}
    \label{fig:interp}
\end{figure*}

\begin{figure}[t]
    \centering
    \includegraphics[width=\linewidth]{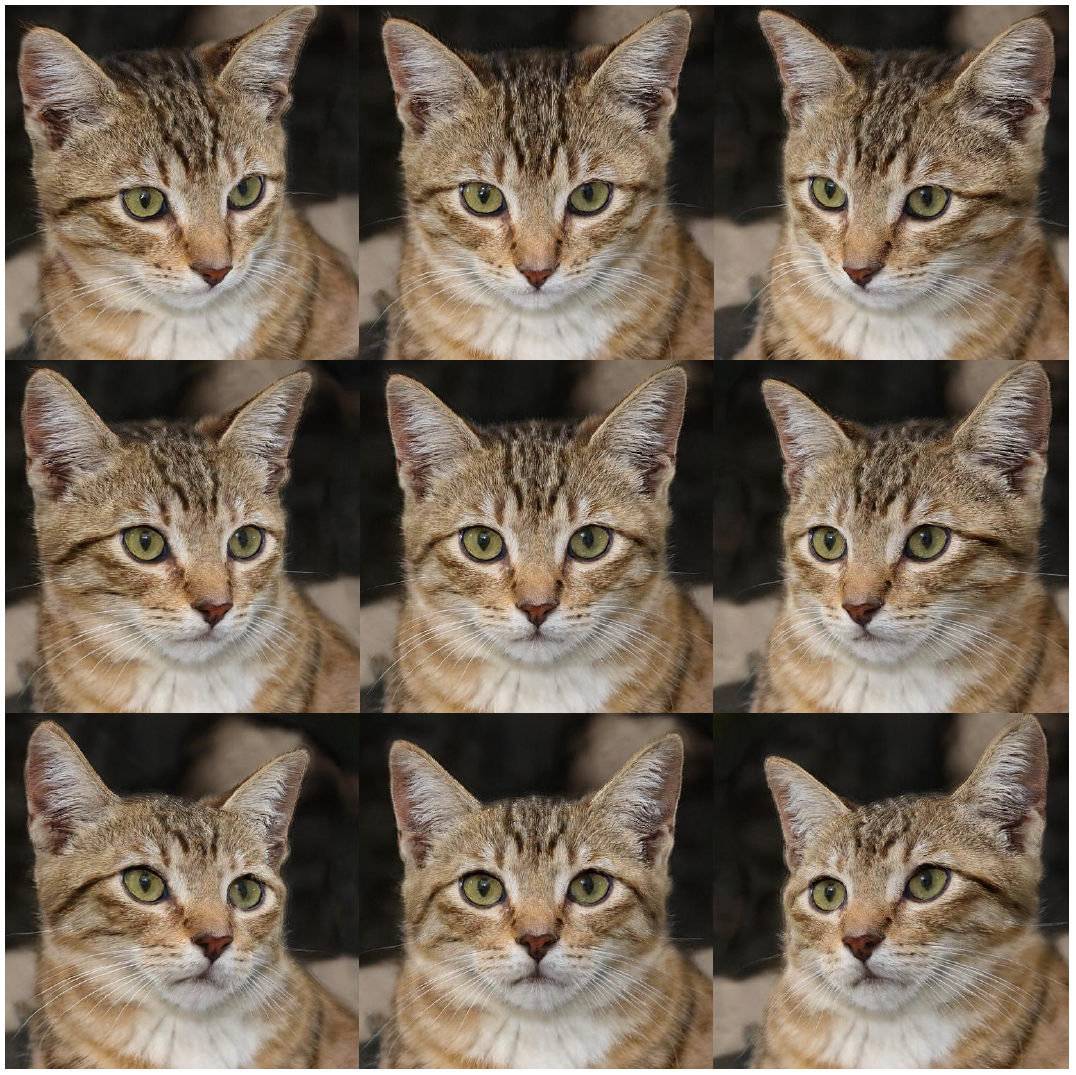}
    \caption{3D-consistent image synthesis. The images synthesized from different camera poses exhibit consistent 3D structures.}
    \label{fig:pose}
\end{figure}

\subsection{Applications}

% \subsubsection{GAN Inversion}
% Our method can be integrated with existing 3D generative models to embed target images into the latent space of our super-resolved model. This embedding is achieved using GAN inversion techniques, which enable the recovery of high-fidelity images at a resolution of $1024^2$, as shown in Fig.~\ref{fig:inv}. Additionally, we explore further applications, such as 3D-consistent image synthesis and style interpolation, which are detailed in the Supplementary Materials.

\subsubsection{3D-Consistent Image Synthesis}
In Fig.~\ref{fig:pose}, we showcase images synthesized from different camera poses. The results clearly demonstrate the high level of 3D-consistency achieved by our method. For an even more comprehensive demonstration of 3D-consistent image synthesis, please refer to our \textbf{Supplementary Video}, which provide dynamic visualizations that further highlight the 3D-consistency of our approach.

\subsubsection{Interpolation in Latent Space}
Fig.~\ref{fig:interp} demonstrates the results of interpolation in latent space. Specifically, given two random latent codes, we interpolate between them and feed the interpolated codes into the generator. This process produces smooth transitions in the synthesized images. Notably, each interpolated result exhibits 3D consistency, demonstrating that images synthesized across the entire latent space can maintain 3D consistency, thanks to our direct rendering strategy.

\subsubsection{High-Resolution Image Synthesis}
\begin{figure*}[t]
  \centering
  \begin{minipage}[t]{\textwidth} 
      \centering
    \includegraphics[width=\linewidth]{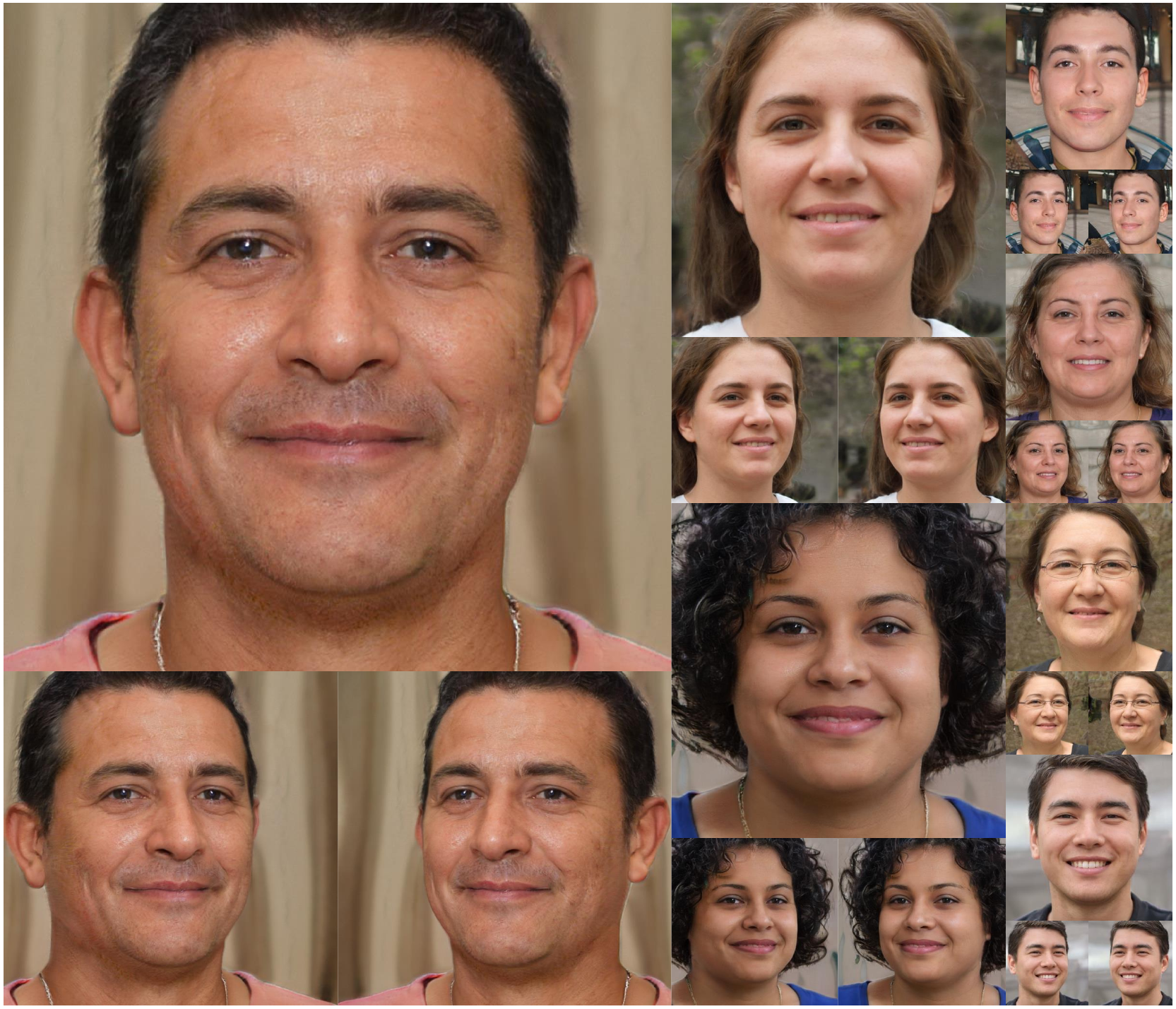}
    \caption{High-resolution ($1024\times1024$) portrait synthesis. Our method produces high-quality portraits with rich details.}
    \label{fig:hr}
  \end{minipage}
  % \begin{minipage}[t]{\textwidth}
  %     \centering
  %   \includegraphics[width=\linewidth]{figure/cat.pdf}
  %   \caption{High-resolution ($512\times512$) cat image synthesis. The resolution is limited by the AFHQ~\cite{choi2020stargan} dataset.}
  %   \label{fig:cat}
  % \end{minipage}
\end{figure*}

Fig.~\ref{fig:hr} showcases high-resolution portraits synthesized by our method. To balance image quality and diversity, we use a truncation technique where the latent code \( w \) is defined as a weighted average of two components: \( w = 0.5 \times w_\mathrm{averaged} + 0.5 \times w_\mathrm{random} \). Here, \( w_\mathrm{averaged} \) is the averaged latent code, ensuring high-quality synthesis, while \( w_\mathrm{random} \) introduces diversity. For each generated entity, we provide images from different viewpoints, demonstrating the 3D-consistency of high-resolution image synthesis—a challenge for previous methods.
% Additionally, Fig.~\ref{fig:cat} presents high-resolution ($512\times512$) cat images, demonstrating our framework's ability to generate detailed and lifelike images across different subject types.

\section{limiation}
Our method achieves stronger 3D consistency than previous approaches by directly rendering high-resolution images. However, this also results in a slight degradation in image quality compared to methods based on 2D super-resolution. Exploring approaches that balance both strong 3D consistency and high image quality would be valuable for broader industrial applications.
Additionally, while our method enables efficient synthesis of high-resolution images through depth-guided rendering, it still requires dense sampling during the low-resolution stage, which can hinder real-time rendering. Replacing NeRF with 3DGS presents a promising alternative, though this area has been explored relatively less.
\section{Conclusion}
In this paper, we present SuperNeRF-GAN, a universal framework for 3D-consistent super-resolution. SuperNeRF-GAN can be seamlessly integrated with existing 3D-aware image synthesis methods to enhance the resolution of synthesized images while maintaining 3D-consistency and efficiency. The core innovation of SuperNeRF-GAN lies in generating a boundary-correct multi-depth map, which is employed in depth-guided rendering to achieve high-resolution image synthesis with enhanced 3D-consistency and efficiency. Compared to existing 3D-consistent methods, our approach consistently demonstrates improvements in both universality and quality. 
% Quantitative comparisons further highlight the superiority of our method in terms of image and geometry quality for high-resolution image synthesis. 
Meanwhile, it is important to note that while our super-resolution framework ensures 3D-consistency, it slightly compromises image quality compared to methods employing 2D image super-resolution.
How to achieve 3D-consistent high resolution image synthesis while ensuring the high image quality similar to the 2D-based methods is still worthy to investigate for realistic image synthesis.
% Thus, achieving high-quality image synthesis with full 3D-consistency remains a significant challenge for realistic image synthesis.

\bibliographystyle{IEEEtran}
\bibliography{ref}

\vfill

\end{document}